# Translating Milli/Microrobots with A Value-Centered Readiness Framework

Hakan Ceylan*[1,2] Edoardo Sinibaldi[3], Sanjay Misra[4], Pankaj J. Pasricha[5], Dietmar W. Hutmacher[2,6-9]

[1] Medical Microrobots Lab, Department of Physiology and Biomedical Engineering, Mayo Clinic, 85259 Scottsdale, Arizona, USA
[2] Max Planck Queensland Centre, Queensland University of Technology, Brisbane, QLD 4000, Australia
[3] Istituto Italiano di Tecnologia, 16163 Genoa, Italy
[4] Department of Radiology, Mayo Clinic, 55902 Rochester, Minnesota, USA
[5] Department of Medicine, Mayo Clinic, 85259 Scottsdale, Arizona, USA
[6] Faculty of Engineering, School of Mechanical, Medical and Process Engineering, Queensland University of Technology, Brisbane, QLD 4000, Australia
[7] Australian Research Council (ARC) Training Centre for Multiscale 3D Imaging, Modelling, and Manufacturing (M3D Innovation), Queensland University of Technology, Brisbane, QLD 4000, Australia
[8] Australian Research Council Training Centre for Cell and Tissue Engineering Technologies, Queensland University of Technology, Brisbane, QLD 4059, Australia
[9] Centre for Behavioral Economics, Society & Technology (BEST), Queensland University of Technology (QUT), Kelvin Grove, QLD, Australia

*Corresponding author's email address: Ceylan.Hakan@mayo.edu

**Abstract**

Untethered mobile milli/microrobots hold transformative potential for interventional medicine by enabling more precise and entirely non-invasive diagnosis and therapy. Realizing this promise requires bridging the gap between groundbreaking laboratory demonstrations and successful clinical integration. Despite remarkable technical progress over the past two decades, most millirobots and microrobots remain confined to laboratory proof-of-concept demonstrations, with limited real-world feasibility. In this Review, we identify key factors that slow translation from bench to bedside, focusing on the disconnect between technical innovation and real-world application. We argue that the long-term impact and sustainability of the field depend on aligning development with unmet medical needs, ensuring applied feasibility, and integrating seamlessly into existing clinical workflows, which are essential pillars for delivering meaningful patient outcomes. To support this shift, we introduce a strategic milli/microrobot Technology Readiness Level framework (mTRL), which maps system development from initial conceptualization to clinical adoption through clearly defined milestones and their associated stepwise activities. The mTRL model provides a structured gauge of technological maturity, a common language for cross-disciplinary collaboration and actionable guidance to accelerate translational development toward new, safer and more efficient interventions.

**Brief Synopsis**

- As small-scale robotics reaches critical mass, the field is entering a new phase where success depends on clinical feasibility and value in addition to technical novelty.
- The translation of milli/microrobots must begin with a well-defined patient need and a rigorous demonstration of safety and efficacy.
- For effective adoption of milli/microrobots, translational strategies should be aligned with seamless workflow integration, such as compatibility with existing interventional imaging systems, surgical suites and hospital practices.
- The proposed milli/microrobot Technology Readiness Level (mTRL) framework aligns engineers, clinicians, regulators, and investors by defining transparent milestones and stepwise activities from establishing feasibility and demonstrating clinical value to adoption.
- Even if not every milli/microrobot reaches the clinic, translational development enriches the collective toolbox, generates regulatory-ready evidence, and builds infrastructure for the next generation of designs.

**Introduction**

Robotic surgery was once a futuristic idea whose time may never come; today it shapes minimally invasive care. Over the past decade, it has enabled greater precision, reduced tissue trauma, and faster recovery(1). Systems such as Intuitive Surgical's Da Vinci have achieved widespread adoption, with over 14 million procedures and more than 76,000 trained surgeons worldwide as of 2025(2). Building on this momentum, newer systems have broadened the field: Mako Spine (Stryker) from orthopedics to, more recently, spine surgery(3); Monarch Platform (Johnson & Johnson) and Galaxy System (Noah Medical) for bronchoscopy(4, 5); and Hugo RAS (Medtronic) emerging as a multi-specialty platform for urology, gynecology and general surgery(6). Their commercial traction and clinical success continue to drive new investment and innovation, setting the stage for a next generation of interventional robotic systems with the potential to deliver even greater technical capabilities and improve patient outcomes(7).

Despite this progress, many clinically critical targets in the vascular system, biliary and reproductive tracts, and fetal or intracranial spaces remain difficult or unsafe to access with current tethered robotic systems and minimally invasive devices. Their elongated, high-aspect ratio designs, relatively bulky end-effectors, and reliance on access ports or line-of-sight visualization impose mechanical and anatomic limitations. Navigating long actuation pathways introduces nonlinear effects such as hysteresis and backlash, while bending or whip-like motions risk unintentional tissue trauma(8, 9). Even minor shear stresses can trigger inflammation or irreversible damage in such sensitive regions(10).

Untethered milli/microrobots can be the next frontier in surgical robotics. Building on the progresses in actuation,

> **Box 1 | Milli/microrobot system**
>
> A milli- or microrobot system is the complete, clinically deployable platform consisting of the miniature robot, which serves as the end effector, together with all essential supporting subsystems. These subsystems include the remote actuation unit (e.g., magnetic, acoustic, or optical control platforms), real-time imaging and tracking modalities (e.g., ultrasound, fluoroscopy), deployment and retrieval devices, and all communication, display, and user-interface software. The integrated "turnkey" system is designed to operate as a cohesive whole within real-world clinical workflows.
>
> For clarity, millirobots are mobile robotic devices that typically operate at the sub-millimeter to millimeter scale ($10^{-2}$-$10^{-4}$ m), where they can be individually tracked with standard imaging and manipulated as discrete tools. Microrobots, by contrast, function at the micrometer to sub-millimeter scale ($10^{-4}$-$10^{-6}$ m), often in swarms or collectives, and are designed for tasks at cellular or microvascular dimensions. Despite their size differences, both require full-system integration for clinical translation.

locomotion, materials, and control over the past two decades, these miniature devices, ranging in size from several millimeters down to the scale of a single cell, can operate as free-floating end effectors inside the human body. Introduced through percutaneous needle punctures or sheaths far smaller than keyhole incisions, they can navigate complex anatomy with minimal or no tissue trauma and reach targets beyond the access limits of conventional systems, opening new diagnostic and therapeutic possibilities. Over two decades of pioneering research has expanded the **toolbox** of small-scale robotics, with rigid and soft/deformable body designs, locomotion modes from swimming to rolling, and functions including biosensing, drug delivery, tissue biopsy, and collective swarm behaviors(11-17). These foundational

> **Box 2 | Toolbox**
>
> In this Review, *toolbox* refers to the growing set of milli/microrobotic capabilities developed through discovery research. These stem from inventive activities in materials, actuation strategies, locomotion modes, control methods, sensing and imaging integration, and fabrication techniques. Collectively, the toolbox enables the design of systems that can be tailored, combined, and matured to meet the demands of specific clinical applications.

efforts have sparked the emergence of startups companies seeking to pioneer entirely new procedures using **milli/microrobot systems**(18-21).

Despite this exciting academic creativity, the transition from laboratory studies to clinical implementation remains slow, fragmented, and immature. Most demonstrations remain proof-of-concept studies under tightly controlled experimental conditions in academic laboratory environments, with limited feasibility for deployment in real-world healthcare settings. In this Review, we identify the root causes of this innovation-to-implementation gap and argue that effective translation requires value-centered research from the outset. This suggests grounding milli/microrobot development on **unmet clinical needs**, practical feasibility, and seamless integration into clinical workflows. To structure this process, we introduce the milli/microrobot Technology Readiness Level (mTRL) framework, which maps system development from concept to clinical adoption through clearly defined milestones and stepwise activities. This roadmap provides actionable guidance for aligning technical innovation with clinical utility, enabling the more effective translation of untethered milli/microrobot technologies from proof of concept to patient impact.

Our intention is not to limit the scope of curiosity-driven milli/microrobot research, but to provide a beacon for a spin-out pathway that can effectively channel growing toolbox capabilities toward real-world viability. Discovery must remain the wellspring of disruptive ideas, providing the freedom to explore uncharted territory without immediate constraints. When the aim is to reach patients, however, the fundamental research must eventually be refined through the filters of regulatory oversight, scalability, usability and eventual adoption.

The Review's timing is strategic. As the field grows and disciplines like materials science and medical imaging extend the multidisciplinary arena of medical robotics, access challenges in interventional medicine are nearing solutions(22, 23). Embedding translational considerations

> **Box 3 | Unmet clinical need**
>
> An *unmet clinical need* is a clearly defined patient problem in which current standard-of-care options fail to deliver acceptable safety, effectiveness, or access for the intended indication The need should be supported by clinical evidence and guideline-based reasoning and framed as a solution-agnostic capability gap. A precise need statement anchors the translational development of milli/microrobots in real clinical practice, not in abstract technical possibility.
>
> **Scenario 1 |** Vascular access
> **Patient:** 68-year-old man with critical limb threatening ischemia; angiography shows heavily calcified, tortuous distal tibial/pedal arteries ($\approx$1.5-2.5 mm). Not a surgical bypass candidate.
> **Problem:** Multiple failed antegrade crossing attempts; standard 0.014–0.018″ wires/microcatheters skate into the calcified plaque in angulated segments, causing intimal injury/perforation risk.
> **Unmet need** A minimally invasive navigation method that maintains true-lumen traversal (or controlled subintimal track with predictable re-entry) through calcified, tortuous distal vessels to develop a definitive endovascular revascularization strategy.
> **Robot design implication:** A device with atraumatic, precise navigation capability in tortuous anatomy; built with hemocompatible, low thrombogenic materials; fluoroscopic visibility; robust retrieval strategies.
>
> **Scenario 2 |** Fetal intervention
> **Patient:** 24-week monochorionic–diamniotic twin pregnancy twin-to-twin transfusion syndrome (Quintero stage II-III) requiring selective ablation of placental vascular anastomoses.
> **Problem:** Current fetoscopes and 8-10 F valved introducers are relatively large/rigid for the chorionic-plate workspace; transabdominal entry through uterus and membranes increases risk of iatrogenic membrane rupture, chorioamniotic separation, and preterm labor.
> **Unmet need:** A low-profile, low-trauma method that is image-guided (ultrasound-first in current practice) to navigate the amniotic cavity, stabilize on the chorionic plate, and deliver vessel-selective ablation with tightly limited thermal spread, minimizing membrane injury and large/rigid hardware.
> **Robot design implication:** Soft and highly maneuverable robot body with size < 2 mm; atraumatic navigation with low contact force/pressure; deployment and retrieval compatible with valved introducer sheaths; image-guidance ready (ultrasound-first); selective vessel ablation capability with limited thermal spread (< 1 mm).
>
> **Scenario 3 |** Gastrointestinal navigation
> **Patient:** 52-year-old woman, prior Roux-en-Y gastric bypass. CT/MRI shows a 1.5–2.0 cm submucosal lesion in the excluded stomach concerning for early gastric cancer.
> **Problem:** Definitive diagnosis requires submucosal tissue. Standard transoral endoscopy cannot reach the remnant; endoscopic ultrasound-assisted access or percutaneous gastrostomy are possible but carry risks (leak, tract creation in a potentially malignant field, staged procedures). Laparoscopic/transgastric surgery entails re-entry in a scarred field with higher morbidity.
> **Unmet need:** A minimally invasive, low-trauma intraluminal method that can navigate altered anatomy to reach the excluded stomach and obtain a depth-controlled submucosal biopsy with hemostasis, while avoiding formal surgery.
> **Robot design implication:** A remotely steerable device with size < 10 mm; atraumatic navigation with traction on wet mucosa and tight turning radius; adjustable biopsy depth to capture > 5 mm mucosal-submucosal specimen; leak-proof specimen containment; safe retrieval strategy.

early through the mTRL framework can help advance milli/microrobotics from labs to practical medical solutions.

**Why Translation Stalls: The Innovation Implementation Gap**

The field of small-scale robotics remains in the "fluid" phase of the innovation life cycle, characterized by rapid design proliferation and conceptual breakthroughs (24). This stage fosters creativity but often comes at the expense of reliability, reproducibility, and scalability. Academic reward structures reinforce this tendency: high publication volume and novelty are more strongly incentivized than durability, system-level integration, or clinical feasibility. As a result, most reports remain proof-of-concept demonstrations under tightly controlled conditions, with limited relevance to real-world practice.

This culture amplifies claims of clinical potential without comparing to existing benchmarks. Unmet clinical needs are often loosely defined, and safety and efficacy relative to current standards remain untested. Essential system-level requirements, such as scalable manufacturing, biocompatibility, sterility, reproducibility, and workflow compatibility, are rarely demonstrated. Fragmentation into specialized sub-domains (materials, actuation, imaging, fabrication) has deepened technical expertise but sidelined integration across the full translational pathway.

The result is a persistent gap between laboratory concepts and clinically implementable solutions, which often leads to inflated expectations and/or misaligned developments. Many milli/microrobots are technically impressive yet "off-target" for medical application: over-engineered for feasibility studies but under-prepared for the practical realities of patient care(24, 25). Bridging this gap requires rethinking how progress is measured, not just by novelty, but by consistent advancement toward clinically defined value.

**Value-Centered Translation of Milli/Microrobots in a Regulated Healthcare Market**

Because the development of medical devices and their indications for use are tightly regulated under national oversight, translational small-scale robotics must be conceived and evaluated with regulation in mind. Effective translation of milli/microrobots begins with a clearly defined unmet clinical need and a framework that prioritizes demonstrable patient value. Technical excellence alone does not guarantee adoption; success requires a systems-level program that anticipates anatomic, physiologic, and procedural constraints, maps regulatory pathways, and aligns with reimbursement and broader healthcare economics. Early, iterative co-development with clinicians, including physicians, nurses, and technical staff, is instrumental for shaping indications, workflow integration, training requirements, and long-term outcomes.

Accordingly, every design choice in a milli/microrobot system must serve a clinically relevant function, contributing to the safety, and effectiveness of the care. Robot size, actuation method, and imaging compatibility must reflect the target anatomy while ensuring safe deployment, retrieval, and operator-friendly use. In practice, barriers such as sterilization, reproducibility, and compliance with hospital standards play the biggest role in determining adoption. Shifting established clinical workflows requires years of cumulative evidence. Empirically, the more complex an innovation, the less likely it is to be adopted, scaled, spread, and sustained in healthcare settings(26).

Progress with the specific milli/microrobot system will also be benchmarked against the standard of care. Regulatory agencies such as the FDA (USA), EMA (EU), and TGA (Australia) typically require evidence of at least non-inferiority, and ideally superiority, in safety and effectiveness relative to existing diagnostics or therapies. Demonstrating credible improvements is critical not only for market authorization, but also for clinician adoption, and, eventually payer reimbursement. Achieving this goal down the road demands establishing feasibility with reproducible validation and demonstrating clinical value in preclinical disease models, regulatory foresight, scalable manufacturing, and system-level completeness.

Without such alignment, even the most inventive milli/microrobots risk remaining confined to the laboratory, stalling in the "valley of death" where promising concepts fail to become viable clinical solutions(27). Innovation studies describe a similar pattern in the Gartner hype cycle: enthusiasm peaks before real-world evidence accrues, often leading to stagnation or collapse. To avoid this fate, small-scale robotics field must embed translational considerations early, when expectations are highest. A value-centered framework anchors robotic engineering to unmet needs of patient, day-to-day care delivery, and the emerging technology's downstream effects on outcomes, costs, staffing, and logistics over years. The reflection of these effects should also be measured transparently and guided insightfully, thereby motivating a dedicated mTRL as a shared yardstick for translational maturity of translational milli/microrobots.

**A Technology Readiness Framework for Milli/Microrobots**

Technology Readiness Level (TRL) scales are widely used across aerospace, automotive, and energy sectors to assess technology maturity (28-30). Over time, TRL has become an objective, evidence-based tool for guiding product development and integrating innovations into complex systems(31). Recognizing the differing demands of each sector, field-specific TRL scales have emerged(32-34). Aligning development plans to TRLs reduces financial and coordination costs, improves time efficiency, and manages risk, which are critical advantages in the regulated medical device market(31). For example, the U.S. Department of Defense, employs comprehensive Technology Readiness Assessment guidelines to control cost growth in new technology initiatives(32).

Small-scale robotics would benefit significantly from a TRL scale formulated explicitly for translational development. A shared, domain-specific rubric would improve communication among cross-disciplinary research teams, clinicians, regulators who may be unfamiliar with the emerging technology's unique features, and investors focused on milli/microrobot systems. Our mTRL framework combines two complementary dimensions: *technical maturity*- evidence that capabilities, such as remote robot actuation, control, imaging compatibility perform in environments of increasing realism, and *clinical alignment*- evidence that those capabilities can create a real clinical value by satisfying a defined medical scenario, including anatomical access, disease indication, procedural constraints, and workflow integration. This dual-axis assessment extends beyond isolated laboratory performance to capture real-world readiness, thereby reducing development risk and directing investment toward applications with the greatest translational potential. It also helps calibrate expectations within the field by highlighting the clinical drivers behind research, the incremental steps required, and discouraging premature or inaccurate claims, which we believe is also essential for public relations.

The mTRL organizes development into four clinical milestone phases comprising nine readiness levels (Figure 1). Table 1 summarizes each level's objectives, typical approaches, and success criteria for assessing the translational maturity of milli/microrobot systems for their intended medical applications. We structure mTRL with the expectation that most milli/microrobot systems will follow a medical device regulatory pathway. Systems whose primary mode of action is metabolic or biochemical may fall under pharmaceutical regulation(33). Some may be classified as combination products that integrate a delivery robotic device with therapeutic cells or drugs.

*Milestone A (mTRL1-4): Identifying unmet clinical need and establishing pre-clinical feasibility*
Milestone A represents the first translational checkpoint. It is reached once a proposed milli/microrobot system has progressed from concept generation (mTRL1) to feasibility demonstration in physiologically relevant models (mTRL4). Achieving this milestone requires iterative refinement of a design concept that carefully addresses unmet clinical need and capable of operating under increasingly realistic conditions.

Translational research starts with the identification of a specific clinical problem and an innovative milli/microrobot concept is developed to address it. At mTRL 1, this is achieved through literature review, initial market surveys, and in-depth discussions with clinicians, surgeons, and other healthcare professionals to understand both the medical need and the landscape of current solutions. The specific aim is to articulate a medically grounded rationale that justifies a milli/microrobot concept that can be potentially superior to existing standard of care.

At mTRL2, the vision develops into a system-level plan. Alternative designs, materials, actuation strategies, and control methods are defined, while physiologic and anatomic constraints are incorporated from the outset. Early design choices must already reflect regulatory foresight and fabrication scalability to support eventual Good Manufacturing Practice (GMP) feasibility, since downstream changes can be both costly and prohibitive for translation. Although prototyping is still performed in phantoms, ex vivo tissue, or small animals, development should increasingly converge toward conditions that simulate an operating room environment.

At mTRL3, controlled non-physiological in vitro experiments (e.g., Petri dish or simplified flow channels) verify that the robot functions reproducibly as a complete system. Iterative prototyping and optimization ensure that all components work seamlessly together, while extensive material and mechanical characterization provides the baseline for safety and performance. Endpoints such as navigation accuracy, imaging visibility, actuation robustness, and nontoxicity are defined here, laying the foundation for subsequent preclinical evaluation.

At mTRL4, testing progresses to physiologically relevant phantoms, ex vivo tissues, and small-animal models. This stage is where robust iteration truly takes hold: performance is validated under increasingly realistic anatomic and pathophysiologic conditions, reproducibility is established, and safety endpoints are refined. Results from this stage form the specific design files and standard operating protocols needed to justify large-animal studies and provide the evidence base for moving higher technology readiness level.

*Milestone B (mTRL5): Demonstrating the proof of clinical value*

While Milestone A focuses mainly on establishing feasibility through iterative refinement of the milli/microrobot system, Milestone B shifts the emphasis to demonstrating clinical value. At this stage, the locked-in lead system design and operational protocols from mTRL4 are evaluated across the full interventional process, asking whether the approach would justify its risks and benefits if applied to patients.

For most researchers in small-scale robotics, the familiar territory ends at mTRL1–4: benchtop proof-of-concepts, phantom models, and small-animal studies. The transition to mTRL5 represents the most significant leap, requiring demonstration of safety, efficacy, and functional value in large-animal disease models that closely replicate human anatomy, pathophysiology, and procedural workflows. These studies provide insight into whether the intervention can alter disease progression, withstand real-world procedural challenges, and generate clinically relevant outcomes within follow-up times comparable to human care.

Achieving this level requires a fully integrated, scalable system and close collaboration with medical teams. Hospital integration plans must also come into sharper focus, covering imaging compatibility, workflow logistics, staff training, and patient safety protocols(34). Importantly, the risk-benefit balance must be made explicit, weighing improvements in safety and efficacy against the economic realities of healthcare delivery.

In practice, achieving Milestone B introduces substantial logistical, ethical, and financial challenges: large-animal housing, surgical support, medical imaging access, and regulatory compliance demand resources. These resources are not typically available at the direct disposal of a single academic lab working on robots. Anticipating these requirements underscores the need for partnerships, with clinicians and clinician scientists, regulatory experts, and industry stakeholders, needed to as to move forward with appropriate resources.

*Milestone C (mTRL6-8): Frist-in-human, clinical trials and regulatory approval*

Once mTRL5 is achieved, the regulatory pathway needs to be clarified in consultation with the relevant authorities, considering the intended use, risk profile, and device classification. In the United States, the 510(k) process allows a device to enter the market by demonstrating substantial equivalence to a legally marketed predicate, often without the need for clinical trials. However, first-generation untethered milli/microrobots are a highly novel device, which will likely require extensive clarification, so the 510(k) pathway would rarely apply. We anticipate that most first-generation milli/microrobots would follow the Class III medical device route, requiring mTRL6 first-in-human trials. These small, tightly controlled studies focus on safety and initial performance, often revealing necessary refinements before broader testing. So far, no human data was shown with milli/microrobots.

As the design variety is diverse in the field of small-scale robotics, in some cases, milli/microrobots may not fit existing classifications. The FDA's De Novo pathway offers an alternative for novel devices that lack a suitable predicate but do not pose the highest risk. For example, in 2020, AnX Robotica's NaviCam™ Magnetically Controlled Capsule Endoscopy System was granted De Novo classification. As autonomy, actuation methods, and therapeutic capabilities evolve, entirely new device categories may be required to address the unique risks and benefits of these systems.

At mTRL7, Phase 2 clinical trials expand to moderate-sized patient cohorts, generating statistically meaningful safety and efficacy data while optimizing procedural protocols and training programs. This stage may also compare the device's performance directly to standard-of-care alternatives.

mTRL8 involves large-scale, multicenter Phase 3 clinical trials to confirm the device's overall risk–benefit profile, establish clinical superiority (or non-inferiority) over conventional treatments, and provide definitive data for regulatory approval and market entry.

*Milestone D (mTRL9): Clinical adoption and continuous improvement*

mTRL9 is reached when the system is commercially available and integrated into routine clinical workflows, supported by reimbursement mechanisms and post-market surveillance. Market approval is not the end of development, nor will it guarantee clinical impact. Minor updates (e.g., software refinements) may occur within mTRL9, while major changes in indication, materials, or fundamental design would require revisiting earlier mTRL stages. Long-term success depends on physician acceptance, ease of training, safety perception, and cost justification. Continuous monitoring of real-world performance ensures that the device evolves in line with clinical needs while maintaining compliance with regulatory and manufacturing standards.

**Enablers of Milli/Microrobot Systems in the mTRL Framework**

This section identifies the cross-cutting enablers that must mature in concert for milli/microrobot translation and maps their maturation in the mTRL roadmap. The core domains are: (i) mechanical robot design; (ii) robotic materials, fabrication, and biocompatibility strategies; (iii) human-scale, image-guided actuation and interventional automation; (iv) imaging, localization, and tracking; (v) deployment, retrieval, and contingency planning; and (vi) standardized methods under a quality management backbone. Across mTRL1-4, the central challenge is progressively integrating these key domains into a coherent, testable system (Figure 2). Because early choices cascade through all later stages, alignment with clinical workflow, regulatory expectations, and usability must be built in from the outset.

*Mechanical Robot Design*

Size is a first-order decision set by the target anatomic site and, in turn, bounds the engineering toolbox—compatible imaging, actuation options, and feasible locomotion(14). Potential areas (nervous, respiratory, urinary, cardiovascular) offer accessible pathways from a few micrometers to a few millimeters. Distal arteries and veins often present ~1–4 mm lumens suited to millirobots, whereas pulmonary microvasculature approaches 2 µm for microrobots. The navigable space in the body is not empty. It is usually fluid-filled, mucus-lined, and the geometric structure can be highly patient-specific. In blood vessels, while arterial cross sections are close to circular, veins are typically elliptical. Vessels are not static structures as described in most of the phantom models used in milli/microrobots. They can deform dynamically in response to pulsation, vasodilation, and vasoconstriction. Consequently, endovascular designs must conform to changing lumen geometry and remain atraumatic to the endothelium, since the preservation of this thin tissue is highly critical for vascular health and inflammation (35). This way a milli/microrobotic drug delivery system can also demonstrate a critical medical superiority over existing devices, such as drug coated balloons for the local management of vascular inflammation- a major unmet need. Propulsion and control (e.g., magnetic torque) must be scaled against competing local tissue forces, such as mechanical tissue compression, viscous drag, and surface friction and verified for feasibility and reliability in more realistic phantom, ex vivo and in vivo testbeds.

*Robotic Materials, Fabrication and Biocompatibility Strategies*

Material selection is one of the earliest and most consequential decisions, as it determines the microfabrication methods, device functionality, performance limits, and biocompatibility risks. Although a broad arsenal exists, including photolithography, electroplating, soft lithography, micromolding, two-photon micro-printing, laser micromachining of foils, thin-film deposition/etching, and micro-assembly, each method imposes constraints on material compatibility, achievable feature size, surface finish, multi-material joining, sterilization tolerance, fabrication throughput and cost(22, 36, 37). Devices must be tested for biocompatibility in their intended application context and planned exposure duration. For example, neodymium-iron-boron (NdFeB) fillers are widely used because they enable strong remanent magnetization, programmable dipoles, and reliable wireless actuation for precise navigation(35, 38). They can be acceptable for temporary interventions if fully encapsulated and leakage-proof. Because NdFeB is corrosion-prone and can release ions, it is generally unsuitable for permanent implantation. Barrier coatings (e.g., parylene C, silicon oxide, Au/Ti) can provide short-term corrosion resistance, but their integrity must be verified under sterilization, mechanical wear, and relevant bodily fluids. Applications involving blood contact additionally require thorough hemocompatibility testing, as mobile microrobots may trigger hemolysis, platelet activation, or thrombosis in ways distinct from conventional devices. These evaluations should follow ISO standards (e.g., ISO 10993-1 for general biocompatibility and ISO 10993-4 for blood interactions)(39). For robots intended for long-term exposure, immunological compatibility becomes critical, since chronic inflammation or foreign body responses can compromise both functionality and patient safety(40). Robot design features such as surface texture, body structure, porosity and degradation behavior must therefore be evaluated case-by-case to minimize adverse responses for the target tissue. If retrieval is uncertain, engineered biodegradation may be considered, but only with controlled fragment size and non-toxic products(41). Radiopaque and echogenic nanoparticles incorporated to the robot body frame provide visibility with fluoroscopic and ultrasound imaging, respectively, and thereby estimating robot pose for reliable locomotion guidance(42). Finally, sterilization compatibility (gamma irradiation, ethylene oxide treatment, low-temperature plasma) must be confirmed without causing demagnetization, swelling, embrittlement and so forth.

*Human-Scale Actuation*

A critical bottleneck for clinical translation lies in human-scale actuation. The overwhelming majority of magnetic milli/microrobots to date have been demonstrated in laboratory settings using benchtop electromagnetic coils or fixed permanent magnets. While these systems provide valuable proof-of-concept control, they do not scale easily to large-animal or human procedures. Recently, a few human-scale magnetic control approaches were reported using coil arrays(43) and robotic permanent magnet systems integrated with open-space imaging modalities(42, 44, 45). These platforms can safely generate magnetic fields in the 5–50 mT range throughout the body, a level sufficient for steering many magnetic milli/microrobots in clinically relevant environments(35).

For translational research, each approach carries distinct trade-offs. Coil-based systems may offer higher flexibility and programmable field control. Still, their complexity, potential cooling challenges in continuous operation, and additional infrastructure costs may create a barrier to widespread adoption in hospital settings. By contrast, permanent magnet systems are more compact and cost-effective. Mounted on robotic arms, permanent magnets can be programmed to move around the patient without disrupting concurrent imaging, providing a pragmatic pathway toward clinical integration. However, moving mechanical parts and large magnet around the patient and other medical equipment might challenge workflow integration. Systems that align with existing surgical suites and integrate seamlessly with imaging equipment are most likely to lower adoption barriers. Within the mTRL framework, human-scale actuation represents a decisive checkpoint in transitioning from laboratory-scale demonstrations to demonstrating potential clinical value.

*Control Methods: The Case for Interventional Automation*

Untethered milli/microrobot systems represent the extreme of surgical miniaturization, where precise actuation, limited onboard sensing, indirect visualization, and nontrivial environmental dynamics (e.g., pulsatile flow, vasospasm, tissue deformation) converge. These characteristics challenge purely manual operation and motivate the use of interventional automation. Milli/microrobots within human body can demand continuous, high-precision input under indirect visualization, such as fluoroscopy. This can raise (i) safety risks from device loss or unintended tissue contact; (ii) longer procedure times; and (iii) cumulative radiation exposure for patients during X-ray-guided interventions (ALARA principles). For tasks such as those requiring complex soft robot kinematic controls, navigation, or swarm dispersion management, computer control can execute repeatable micromotions faster and more reliable, enforce hard regional safety limits within the tissue, reduce total procedure time, patient exposure to radiation and the operator's cognitive load. The interventionalist retains live monitoring with instant override. The level of autonomy should be risk-based, conditioned by anatomy, physiology, imaging latency, and robot design-ranging from fully manual through assistive/shared control to supervised task autonomy and, only when validated, full autonomy(46).

*Imaging, Localization and Tracking*

Seamless compatibility with FDA-approved, widely available imaging modalities already present in hospitals, such as fluoroscopy and ultrasound, is strategic for translational progress. Leveraging existing platforms ensures workflow feasibility and avoids the prohibitive costs of building entirely new infrastructures. Nevertheless, major technological gaps remain: no single modality currently provides the ideal combination of fast acquisition, deep tissue penetration, and high-resolution anatomical context needed for the safe real-time tracking and steering of milli/microrobots. Fluoroscopy can visualize robots across a wide size range (0.1–10 mm), while lacking detailed anatomical context and three-dimensional localization. Therefore, safe navigation remains limited. Ultrasound struggles in regions near bone or air cavities, limiting many intracranial and gastrointestinal applications.

　　Translational research should focus on developing new approaches to alleviate such technical gaps for effective in-vivo robot control. A recent emerging concept to this end has proposed integrating fluoroscopic guidance with a virtual reality (VR) environment that houses a digital twin of the operational workspace and a robot avatar(42). This visual enhancement approach allows robust tracking and steering with minimal latency, while providing clinicians with an augmented anatomical view. Such frameworks could also enable the integration of additional intraoperative data streams, such as robot velocity, blood flow dynamics, or periodic organ motion, into real-time control, potentially improving both safety and precision. While promising, these innovations themselves must still be judged against translational criteria: interoperability with existing clinical workflows, reproducibility in physiologically relevant settings, and regulatory acceptance. Within the mTRL framework, such concepts exemplify how emerging technologies can accelerate feasibility if they are developed not as stand-alone research demonstrations but as integrated, workflow-compatible solutions.

*Deployment, Retrieval and Contingency Strategies*

Deployment and retrieval remain underexplored but decisive for clinical adoption. For endovascular robots, delivery catheters can facilitate controlled deployment, while retrieval may involve reversing the robot into the catheter or employing magnetic retrieval catheters(47). When retrieval is not feasible, biodegradable designs provide an alternative path, ensuring safe elimination from the body and reducing long-term risks, such as inflammation or fibrosis (41). Each strategy entails trade-offs in terms of safety, feasibility, and regulatory acceptance, which must be aligned with the intended application. For example, while biodegradability can mitigate chronic exposure risks, in vascular settings it may also increase the danger of embolization, requiring careful material and design choices.

*Quality Management System for the Standardization of Methods*

Another decisive step moving beyond Milestone B is complying with the increased volume and precision standards of documentation. A Quality Management System (QMS) or equivalent structured framework provides standardized, auditable processes across the product lifecycle, from design and prototyping to manufacturing and testing. Frameworks such as ISO 13485 or the FDA's Quality System Regulation form the basis for regulatory credibility. While full QMS implementation is mandatory for companies, adopting QMS-aligned practices in academic translational research can prevent costly repetition and streamline the path to clinical trials.

**Economics of Translation**

Translational innovation is at least an order of magnitude more expensive than discovery research. In the United States, developing a novel Class III therapeutic device from concept to FDA approval typically costs $54 million, with costs ranging from $25 million to $200 million (48). For milli/microrobots, these financial demands accumulate gradually across multiple technology readiness levels (mTRLs), with distinct funding models and risk profiles at each stage.

At early mTRL levels (mTRL1-5), costs are driven primarily by prototyping, benchtop validation, and small animal studies. Funding typically comes from academic grants (e.g., NIH R01 in the U.S., ERC in Europe, NHMRC Investigator/ARC Discovery in Australia, KAKENHI in Japan, or NSFC General Program in China), philanthropic foundations, or seed accelerators, since devices remain afar from regulatory pathways. Prototypes are often artisanal, with no requirement for manufacturing to regulatory-grade standards. However, even at this stage, system-level complexity adds costs: a microrobot platform is not just the robot itself, but also requires actuation systems, imaging integration, and real-time localization, tracking and navigation software. Each interface brings regulatory hurdles, as medical software standards must be followed.

A decisive inflection point where costs rise sharply would come after mTRL5 toward "first in human" readiness in mTRL6 activities. At this point, similar to other medical device translation, milli/microrobots must also transition to Good Manufacturing Practice (GMP) production, undergo Good Laboratory Practices (GLP) testing, and engage with regulators through FDA pre-submissions or Investigational Device Exemption applications. Academic funding rarely covers these activities, making capital investment essential. Typical sources include early-stage venture capital, translational research funds, or strategic industry partnerships. A major expense here is establishing reproducible GMP manufacturing, requiring consistent material properties, validated mechanical/magnetic performance, and robust sterilization protocols.

Costs intensify with first-in-human feasibility trials, requiring hospital integration, Institutional Review Board milli/microrobots approvals, and clinical-grade quality systems (ISO 13485). At this stage, hidden costs emerge from workflow disruption: additional procedural time, staff training, and possible redesign of operating room infrastructure. Funding often comes from venture capital, government translational programs, or public–private consortia.

In medical device development, the largest financial burden often comes with multicenter clinical trials and large-scale GMP manufacturing. Here, industry partnerships, late-stage venture capital, or corporate acquisitions typically drive funding. Success depends as much on payer acceptance and cost-effectiveness as on safety and efficacy, since milli/microrobots will be compared against closest alternatives such as catheters, stents, or tethered robots. Demonstrating clear economic value, superiority or non-inferiority at lower cost, is critical for both reimbursement and ultimately clinical adoption.

**Key Lessons from Other Transformative Medical Device Adoption**

The history of successful medical devices challenges the often-attributed Henry Ford quotation, "If I had asked people what they wanted, they would have said faster horses". In healthcare, translation is a value pull rather than a technology push, meaning that adoption follows when innovations deliver demonstrable patient benefits, integrate into clinical practice, and build trust. Transformative technologies such as coronary stents, pacemakers, cochlear implants and Da Vinci robot were not adopted merely because they were disruptive; they succeeded because they solved big unmet clinical needs in reliable and measurable ways, the result of years of focused, systematic, and iterative development around clearly defined problems.

The history of such medical device innovation offers valuable insights into the translational trajectory of milli/microrobots. Landmark cases reveal recurring themes that determine whether what is often described in publications as technically novel devices may ultimately succeed in clinical practice.

*Workflow integration is decisive*

The widespread adoption of coronary stents was driven not only by their ability to reduce restenosis but also by their seamless incorporation into a catheter-based infrastructure that had been developed over decades, spanning angiography, fluoroscopic imaging, and balloon angioplasty since the late 1970s (49). Because stents could be deployed using the same interventional tools and techniques already familiar to physicians, they required no radical reorganization of practice. This continuity greatly accelerated their acceptance in cardiology suites along with the randomized clinical trials in 1994 that demonstrated their clinical superiority(50, 51).

*Reliability outweighs novelty*

Early pacemakers, introduced in the late 1950s, demonstrated the feasibility of long-term cardiac pacing; however, frequent device failures, often due to battery depletion or lead malfunctions, eroded clinical trust. Over the following decades, incremental progress and iterative refinements in power supply, hermetic sealing, and lead design transformed these fragile prototypes into robust, reproducible devices. Only when reliability was achieved did widespread clinical adoption would follow. For milli/microrobots, establishing feasibility and demonstrating clinical value will likely follow a similar arc of incremental improvements that gradually establish clinical trust.

*Clear patient benefit ultimately drives acceptance*

Adoption of cochlear implants illustrates how even transformative technologies can face deep scepticism before adoption. When first introduced in the late 1970s and 1980s, many clinicians argued that the few electrodes in early devices could never reproduce intelligible speech, regulators worried about safety, and the Deaf community resisted what was perceived as an attempt to "cure" deafness(52). For more than a decade, implants were dismissed as experimental. Only when long-term studies demonstrated that multi-channel devices restored meaningful auditory function, enabling speech perception and language development, did perceptions shift and cochlear implants become established as the standard of care.

*Adoption requires an ecosystem*

The Da Vinci surgical system shows that adoption requires more than technical capability; it needs an ecosystem. When first introduced in the late 1990s, the platform faced widespread skepticism:

its cost was prohibitive, many surgeons doubted it offered meaningful advantages over established minimally invasive techniques, and hospitals hesitated to invest in a device with uncertain reimbursement. What ultimately shifted perceptions was not just the robot's dexterity, but the company's creation of a supportive ecosystem: standardized surgeon training programs, on-site technical support, service contracts that guaranteed uptime, and aggressive marketing that framed robotic surgery as cutting-edge. These non-technical factors, combined with gradual clinical evidence, created a pathway for integration, transforming Da Vinci from a costly curiosity into a dominant surgical robotics platform.

**CONCLUSION**

As milli/microrobotics gathers critical mass, the field is entering a phase defined by translational development. This phase emphasizes addressing real-world clinical challenges and advancing the technological readiness of robotic systems for medical use. It marks a shift from the dominant focus on discovery and proof-of-concept research toward deliberate, stepwise progress aimed at solving well-defined unmet medical needs.

From an academic perspective, while successfully advancing through the mTRL stages cannot guarantee clinical or commercial success, it can significantly increase the likelihood of achieving a lasting medical impact. Even when market adoption proves slower or more limited than expected, the translational process yields tangible benefits. It sharpens engineering practice, produces validated preclinical and clinical protocols, and generates regulatory-ready knowledge that future projects can build further upon. In this way, advancing a milli/microrobot design along a structured technology readiness pathway is not a binary proposition of success or failure, but an investment in a growing technological toolbox for the community. The challenges of translation should not deter researchers; instead, they should view it as a natural extension of fundamental discovery, with the potential to transform small-scale robotics into meaningful clinical tools aimed at advancing healthcare and improving the quality of life of people on a global scale.

**Acknowledgement**

Funding: H.C. acknowledges financial support for this study from Mayo Clinic startup funding; American Heart Association Career Development Award (award number 23CDA1040585); Career Development Award in Cardiovascular Disease Research Honoring Dr. Earl W. Wood, administered by the Mayo Clinic Center for Clinical and Translational Science; the State of Arizona, Arizona Biomedical Research Centre New Innovator Award (award number RFGA2024-022-002); and the National Heart, Lung, and Blood Institute of the National Institutes of Health under Award Number R01HL179093). DWH acknowledge support from the Max Planck

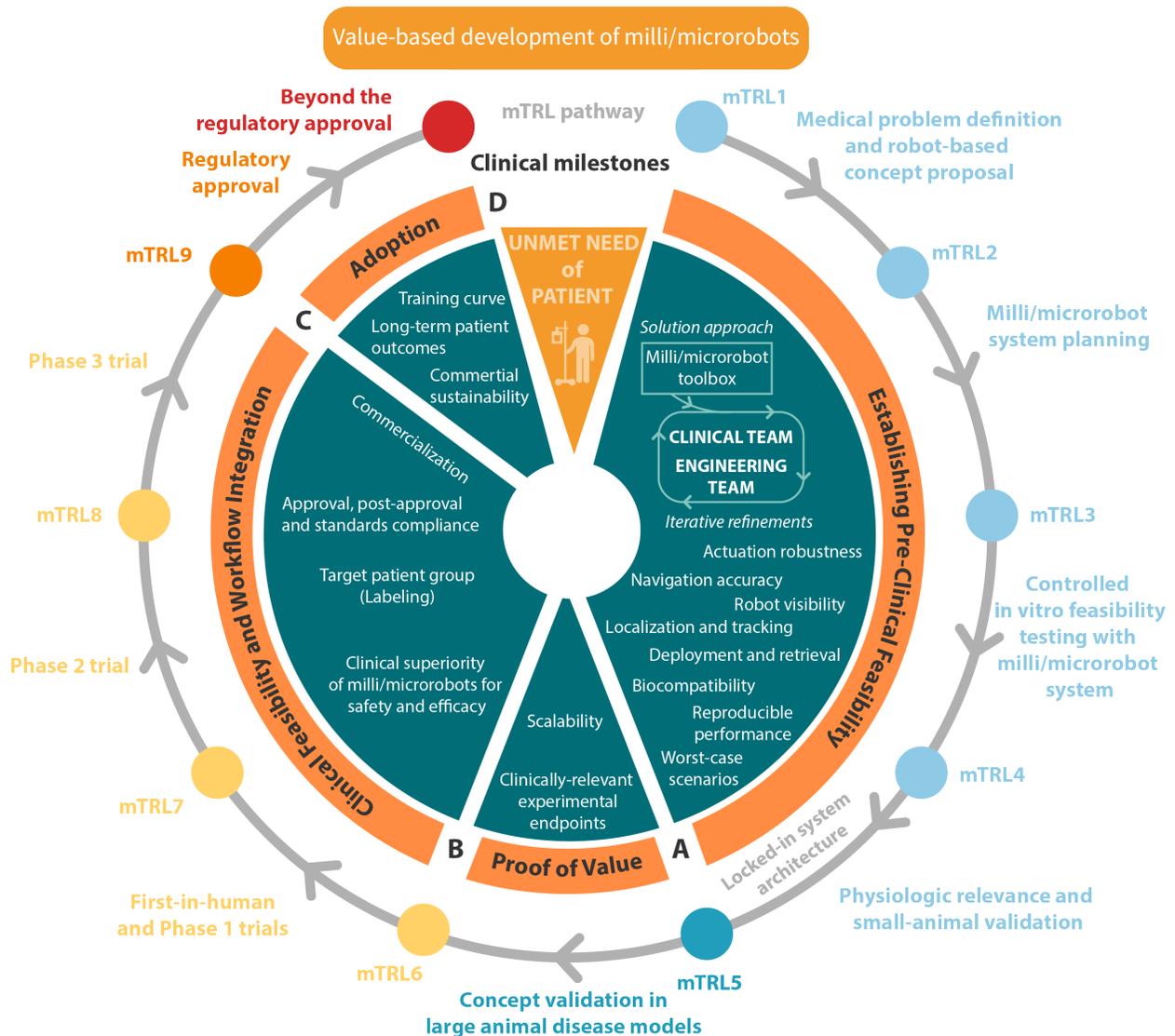

**Figure 1. Value-based translation roadmap of untethered milli/microrobots.** Translating milli/microrobots from proof-of-concept into the clinic requires aligning innovation with unmet patient needs, feasibility, and workflow integration within a systems-level, value-centered roadmap. Translation unfolds iteratively along the milli/microrobot Technology Readiness Level (mTRL) pathway, from unmet clinical need (mTRL1) to commercialization and post-market surveillance (mTRL9). Progress relies on continuous interaction among five key stakeholders: the clinical team (defining unmet needs, ensuring workflow integration and adoption), patients (shaping outcome priorities and managing perceptions critical for acceptance), engineers (developing physiologically and procedurally compatible systems), regulatory bodies (ensuring safety and compliance), and market stakeholders (securing reimbursement, adoption, and sustainability). Together, these domains form a dynamic feedback loop where insights at each stage refine earlier steps, maximizing the likelihood that innovations remain clinically relevant, adoptable, and sustainable.

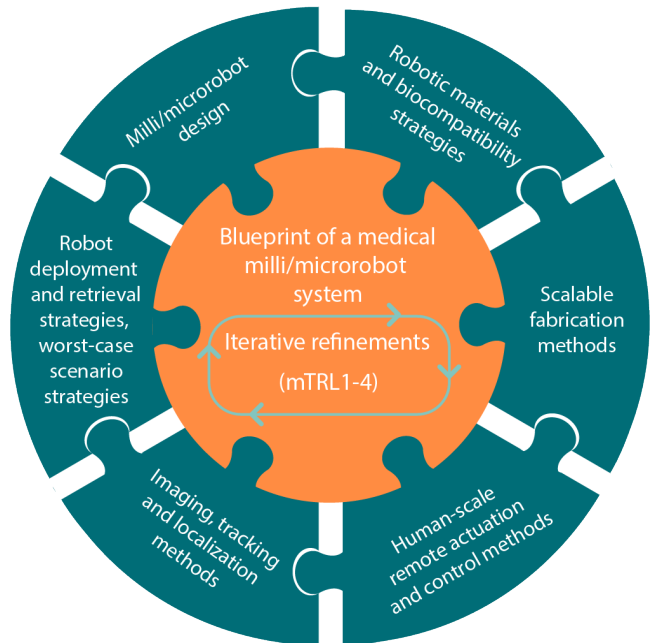

**Figure 2. Creating the early blueprint of a medical milli/microrobot system is akin to solving a puzzle.** Core elements, robot design, materials, fabrication, actuation, imaging, deployment, retrieval, and contingency strategies must interlock into a coherent whole tailored to a specific anatomic and pathophysiologic setting. Once assembled for a given application, flexibility for major changes becomes limited. Even seemingly small modifications (e.g., imaging modality, material composition) may cascade across mTRLs1-4, requiring redesign and revalidation. This interdependence highlights why early design alignment with clinical workflows and regulatory foresight is essential for translational progress.

**Table 1. Technology Readiness Level (mTRL) pathway for milli/microrobots**

| mTRL | Description | Key Activities |
|---|---|---|
| 1 | **Unmet clinical need defined** | • Objective: Identify a clinically significant unmet need and propose an innovative intervention strategy using a milli/microrobot approach.<br>• Approach: Conduct literature reviews, initial market and regulatory landscape scans, and stakeholder interviews with healthcare providers (physicians, surgeons, nurses, biomedical engineers, technicians) to map current practice and limitations.<br>• Success criteria: Documented unmet need with evidence-based justification, and preliminary assessment of potential advantages over existing solutions. |
| 2 | **Holistic plan developed** | • Objective: Formulate a multidisciplinary project plan integrating clinical, engineering, and regulatory perspectives.<br>• Approach: Assemble a multi- and cross-disciplinary teams; outline candidate system architectures, materials, actuation, control, sensing/imaging strategies, fabrication methods, and testing approaches; consider anatomical and physiological constraints of the target site. Identify worst-case scenarios and safety risks.<br>• Success criteria: Defined system architecture and development plan that is technically plausible, clinically relevant, and aligned with regulatory and manufacturing constraints. |
| 3 | **Proof-of-concept demonstrated in non-physiologic testbeds** | • Objective: Demonstrate initial proof-of-concept demonstrated in controlled lab environment.<br>• Approach: Hypothesis-driven experiments using benchtop or phantom testbeds; early physical/chemical characterization; iterative prototyping; initial integration of key subsystems; preliminary biocompatibility screening; early risk assessment for safety-critical features.<br>• Success criteria: Reproducible performance in early test environments meeting initial design targets; all design requirements and rationale are documented with clearly defined and measurable endpoints proposed for subsequent preclinical animal studies. |
| 4 | **Feasibility established in physiologically** | • Objective: Establish feasibility and refine performance under anatomically and physiologically realistic conditions. |

| | relevant phantom, ex vivo and in vivo models | - Approach: Testing in realistic phantoms, ex vivo tissues, and small animal models; iteration toward clinically relevant scale and function; definition of assays, surrogate markers, and endpoints for later non-clinical and clinical studies; initial workflow integration with imaging/actuation platforms.<br>- Success criteria: Demonstrated navigation, control, and function under realistic conditions; initiation of Quality Management System documentation (technical file, design history file); draft regulatory strategy. |
|---|---|---|
| 5 | **Proof of value demonstrated in a large animal disease model** | - Objective: Demonstrate therapeutic/diagnostic benefit and safety compared with current standards in large animal disease models.<br>- Approach: Preclinical studies in disease-relevant large animals with clinically appropriate follow-up; scale-up of actuation/imaging hardware to clinical ranges; refinement of clinical workflows with practitioner input.<br>- Success criteria: Finalized ("locked") system design; complete design history file; positive comparative performance data; regulatory consultations completed; pre-Investigational Device Exemption (IDE) for FDA or equivalent submission filed; draft product development plan. |
| 6 | **First-in-human and phase 1 clinical trials** | - Objective: Assess safety, usability, and initial performance in humans.<br>- Approach: Secure regulatory approval for human use; conduct first-in-human studies, followed by Phase 1 trials in a small patient cohort; collect feasibility, safety and preliminary efficacy data.<br>- Success criteria: Acceptable safety profile; functional performance consistent with preclinical results; data package adequate to design Phase 2 trial. |
| 7 | **Phase 2 clinical trial** | - Objective: Gather statistically valid evidence of safety and preliminary clinical efficacy in the intended patient population, refining device use parameters and procedural protocols.<br>- Approach: Conduct a controlled clinical study in a moderate-sized cohort that represents the target indication, with endpoints defined in consultation with regulators. Collect data on clinical performance, usability, patient outcomes, and any device-related adverse events. |
| | | • Success criteria: Sufficient safety and efficacy evidence to justify progression to a large-scale pivotal (Phase 3) trial; validated procedural protocols and refined device specifications for final verification. |
| 8 | **Phase 3 clinical trial** | • Objective: Evaluate the overall risk-benefit of the therapy and to provide a basis for product labelling.<br>• Approach: Large-scale trials comparing the new intervention to standard-of-care treatments.<br>• Success criteria: Demonstrated clinical superiority, non-inferiority, or added value over conventional treatments; data sufficient for regulatory approval and market entry. |
| 9 | **Commercialization and post marketing surveillance** | • Objective: Launch the milli/microrobot system as a medical device, monitor long-term safety and effectiveness, and ensure sustained compliance.<br>• Approach: Manufacturing under Good Manufacturing Practice (GMP); market rollout; clinician training; adverse event reporting; periodic safety updates; minor or major product iterations as needed.<br>• Success criteria: Continued regulatory compliance; acceptable safety profile in real-world use; reimbursement pathways established; commercial sustainability achieved. |